\documentclass[runningheads]{llncs}
\usepackage{graphicx}

\usepackage{tikz}
\usepackage{comment} 
\usepackage{amsmath,amssymb} 
\usepackage{color}
\usepackage[labelfont=bf]{caption}
\usepackage[utf8]{inputenc}
\usepackage{multirow}
\usepackage{multicol}
\usepackage{amsmath}
\usepackage{amsfonts}
\usepackage{float}
\usepackage{tabularx}
\usepackage{makecell}
\usepackage{subcaption}
\usepackage[font={footnotesize}]{caption}
\usepackage{threeparttable}
\usepackage{url}
\usepackage{hyperref}
\usepackage{cite}

\xdefinecolor{tucol1}{rgb}{1.0 0.5 0.05}
\xdefinecolor{tucol2}{rgb}{0.52,0.72,0.10}
\xdefinecolor{tucol3}{rgb}{0.0 0.0 0.8}
\xdefinecolor{tucol4}{rgb}{0.8 0.8 0}
\xdefinecolor{tucolv}{rgb}{0.1 0.35 0.17}
\xdefinecolor{tucol5}{rgb}{0.8 0 0.8}
\xdefinecolor{tucol6}{rgb}{0 0.8 0.8}
\xdefinecolor{tucol7}{rgb}{0.7 0.8 0.1}
\xdefinecolor{tucol8}{rgb}{0.2 0.7 0.34}
\xdefinecolor{tucol9}{rgb}{0.7 0.8 0.3}
\xdefinecolor{tucol10}{rgb}{0.5 0.5 0.5}
\xdefinecolor{tucol11}{rgb}{0.9 0.9 0.3}
\xdefinecolor{blue}{rgb}{0.4 0.5 1}
\xdefinecolor{softgreen}{rgb}{0.49,0.50,0.47}

\begin{document}
\pagestyle{headings}
\mainmatter

\title{Multi-Channel Time-Series Person and Soft-Biometric Identification} 

\titlerunning{Multi-Channel Time-Series Person and Soft-Biometric Identification}
%
\author{Nilah Ravi Nair\inst{1}\orcidID{0000-0001-9576-0296} \and \\
Fernando Moya Rueda \inst{2}\orcidID{0000-0001-8115-5968} \and \\
Christopher Reining\inst{1}\orcidID{0000-0003-4915-4070} \and \\ Gernot A. Fink \inst{2}\orcidID{0000-0002-7446-7813}}
\authorrunning{N. Nair et al.}
%
\institute{Chair of Material Handling and Warehousing \and
Pattern Recognition in Embedded Systems Group\\ TU Dortmund University, Dortmund, Germany\\
\email{nilah.nair,fernando.moya,christopher.reining,gernot.fink@tu-dortmund.de}}

\maketitle

\begin{abstract}
Multi-channel time-series datasets are popular in the context of human activity recognition (HAR). On-body device (OBD) recordings of human movements are often preferred for HAR applications not only for their reliability but as an approach for identity protection, e.g., in industrial settings. Contradictory, the gait activity is a biometric, as the cyclic movement is distinctive and collectable. In addition, the gait cycle has proven to contain soft-biometric information of human groups, such as age and height. Though general human movements have not been considered a biometric, they might contain identity information. This work investigates person and soft-biometrics identification from OBD recordings of humans performing different activities using deep architectures. Furthermore, we propose the use of attribute representation for soft-biometric identification. We evaluate the method on four datasets of multi-channel time-series HAR, measuring the performance of a person and soft-biometrics identification and its relation concerning performed activities. We find that person identification is not limited to gait activity. The impact of activities on the identification performance was found to be training and dataset specific. Soft-biometric based attribute representation shows promising results and emphasis the necessity of larger datasets. 

\end{abstract}

\section{Introduction}

Biometrics are physical or behavioural characteristics that can be used to identify an individual \cite{jainIntroductionBiometricRecognition2004}. These characteristics need to be universal, distinct, permanent and collectable. For example, fingerprint, signature and gait. Biometrics have gained importance due to security applications. For instance, security footage of individuals in gait motion can be used for their recognition and authentication. However, variation in clothing, multiple subjects in the frame, occlusion and obtrusion are challenges to vision-based gait-based person identification \cite{liuSurveyGaitRecognition2009}. Using on-body devices (OBDs) can pose a solution to these problems. OBDs, such as smartphones or smartwatches, record gait motion that can facilitate recognition of the individual \cite{goharPersonReIdentificationUsing2020}. Furthermore, OBD recordings can be used to classify individuals based on soft-biometrics, such as age, gender identity and height as given by \cite{riazOneSmallStep2015}. 
    
HAR, using OBDs, is highly researched due to its applications in industrial environments, clinical diagnosis and activity of daily living (ADL) monitoring \cite{moyaruedaConvolutionalNeuralNetworks2018, niemannLARaCreatingDataset2020, ruedaHumanPoseOnBody2021}. Real human activity recordings are often required to facilitate process optimisation or real-time HAR. Consequently, the possibility of person identification and soft-biometrics-based classification lead to privacy concerns. \cite{goharPersonReIdentificationUsing2020} raised privacy concerns based on gait-based person identification. \cite{elkaderWearableSensorsRecognizing2018} concluded that OBD-based identification is not limited to gait; activities such as laying, running, and cleaning can contain an individual's identity. Person identification is particularly interesting for biometric authentication on smartphones \cite{mekruksavanichBiometricUserIdentification2021}.
    
Research on bench-marked datasets is limited due to the unavailability of large, annotated datasets with varying human characteristics and natural body movements. Furthermore, given the case of OBD sensors, the influence of sensor biases needs to be analysed. As a result, this paper serves as a preliminary step towards creating efficient identity protection methods, benchmarked research and motivation for creating large annotated datasets. Thus, we experiment on four benchmark OBD datasets intended for HAR in the context of person identification. Datasets with natural body-movements were preferred. The impact of the activity on identity relations is analysed. Furthermore, the possibility to model soft-biometrics as attribute representation is examined. To avoid manual feature extraction \cite{dehzangiIMUBasedGaitRecognition2017}, experiments will be performed on Neural Networks, such as Convolutional Neural Networks (CNNs) and Recurrent Neural Networks (RNNs). For the concept of identity protection, methods to either mask or delete identity need to be researched. This process may bring to light the effect of soft-biometrics and the subject’s individuality on the HAR dataset and thus, the possibility of enhancing HAR dataset. Explainable Artificial Intelligence (XAI) method is explored to investigate its feasibility.

\section{Related Work}

Using gait as a biometric is challenging due to its sensitivity to changes in clothing, fatigue and environmental factors \cite{boydBiometricGaitRecognition2005}. Creating a database template for verification and identification with consideration of these variations is cumbersome. However, gait is one of the most collectable data alongside signature, face thermogram, and hand geometry \cite{jainIntroductionBiometricRecognition2004}. The authors in \cite{singhVisionBasedGaitRecognition2018} presents a survey of the various methods of person identification using low-resolution video recordings. Silhouette analysis \cite{liangwangSilhouetteAnalysisbasedGait2003} and Gait Energy Image (GEI) \cite{hanIndividualRecognitionUsing2006} are two popular video-based gait-based person identification methods. 
   
Based on gait data acquired from OBDs, \cite{shahidStudyHumanGait2012} mention that the gait parameter study can be extended to person identification and soft-biometrics. The authors in \cite{riazOneSmallStep2015} confirmed that a single step recorded from smartphones or smartwatches could be used to reveal soft-biometrics. 

Due to the susceptibility of manual feature extraction to subjectivity, \cite{dehzangiIMUBasedGaitRecognition2017} performed gait-based person identification using multi-layer sensor fusion-based deep CNN. Furthermore, the late fusion method achieves better accuracy, given a defective OBD amongst a group of sensors. In contrast, the authors in \cite{chunshengHumanGaitFeature2020} performed gait-based person identification using a simple Neural Network. The authors found that variations in the gait characteristics are based on the subject's height, weight, arm length and personal habits. The authors in \cite{goharPersonReIdentificationUsing2020} attempted person identification on Gated Recurrent Units (GRU) and Long-Short Term Memory (LSTM). The authors concluded that the GRU model performs better than CNN and LSTM. Besides, they identified that the \textit{step} data performed better than the \textit{stride} data. A \textit{step} is defined as the heel strike of one foot followed by the heel strike of the other foot. \textit{Stride} constitutes two heel strikes from one foot. During a typical human walk, the step frequency is between $1$-$2Hz$. 

In contrast to the previous works, the authors of \cite{elkaderWearableSensorsRecognizing2018} considered a dataset of $20$ daily human activities, such as cleaning, washing dishes, and office-work activities, to perform person identification. Unlike \cite{dehzangiIMUBasedGaitRecognition2017}, the authors focused on manually extracted features, such as mean, standard deviations (SD) and magnitude. The experiments were performed on classifiers such as Support Vector Machines (SVM), K-Nearest Neighbour (KNN), Neural Network, Decision Tree (DT) and their types. The authors identified that sedentary activities had a higher classification rate. In addition, the authors concluded that all subjects are not equally identifiable. The conclusions derived can be pivotal for research in identity anonymization as found in \cite{malekzadehMobileSensorData2019}. Here, the authors used a Convolutional Auto-Encoder to remove the identity information present in the OBD data obtained from a smartphone while maintaining application-specific data, for example, activity recognition and step count. The authors achieved $92\%$ accuracy on activity recognition while reducing user identification accuracy to less than $7\%$. However, the authors have not considered motionless activities based on their observation that subjects can only be distinguished based on motion activities. 
    
The authors in \cite{retsinasPersonIdentificationUsing2020} experimented on person identification using a Deep Neural Network (DNN) and the impact of the sensor noise on the identification. The dataset used consisted of $20$ individuals and was annotated with three activity labels, namely, sleep, walk and other. They identified that sleep activity provided the worst person identification, whereas walking activity performed the best. Further, the authors noted that, noise removal methods using Short-Time Fourier Transform (STFT) does not solve sensor dependency issue. However, raw data augmentation helps with achieving sensor independent network. Nevertheless, person identification was achieved at a lesser accuracy rate. \cite{mekruksavanichBiometricUserIdentification2021} experimented with smartphone data UCI-HAR and USC-HAD datasets on CNN, LSTM, CNN-LSTM and ConvLSTM and was able to achieve high accuracy levels for all users. They implemented a sequence of classifiers to perform activity classification followed by user identification. The authors targeted biometric user identification based on mobile platforms. These experiments emphasize the importance of large, annotated datasets with varying human characteristics and sensor noise analysis.

The authors of \cite{niemannLARaCreatingDataset2020, reiningAnnotationPerformanceMultichannel2020} created the Logistic Activity Recognition Challenge (LARa) dataset and annotated the data based on activity classes and attribute representations. Attribute representations describe coarse objects and scenes, as explained in \cite{farhadiDescribingObjectsTheir2009, lampertAttributeBasedClassificationZeroShot2014}, and activity classes, as explored in \cite{ruedaLearningAttributeRepresentation2018a, reiningAttributeRepresentationHuman2018}. For HAR, linking the body part movement to the activity class helps to describe the class better as well as navigate inter and intra class misclassifications. Given that an individual can be described based on soft-biometrics, it is of interest to extend the idea to person identification by modelling soft-biometrics as attribute representation. 
  
Given the nascent stage of research and variation in experimental approach, person identification using OBDs have conflicting findings. Thus, one needs to experiment on publicly available datasets to ensure reproducible experiments. To investigate the impact of noise and activity, using Explainable Artificial Intelligence methods (XAI) may help to verify the observations. For example, \cite{horstExplainingUniqueNature2019} experiments on ground reaction forces (GRF) verified that the individual's gait pattern have unique characteristics based on the kinematic and kinetic variables. To verify, the authors used Layer-Wise Relevance Propagation (LRP) \cite{montavonLayerWiseRelevancePropagation2019}, a method of XAI. Similarly, LRP method can be used to verify the results of person identification using motion information. Research in this direction may show feasibility of noise to signal segregation.

\section{Person and Soft-Biometrics Identification}

\begin{table}[ht]
\caption{Attribute representations are created based on the recording protocol of LARa dataset.}%
\resizebox{\columnwidth}{!}{
\begin{tabular}{c|ccccc|cccc}
    & \multicolumn{5}{c|}{\textbf{Recording Protocol}}            & \multicolumn{4}{c}{\textbf{Attribute Representation A\_1}}                                                                                                                                        \\
\textbf{Sub} & \textbf{Gender}   & \textbf{Age} & \textbf{Weight}   & \textbf{Height}   & \textbf{Handedness} & \textbf{Gender}    & \textbf{Age}                                                     & \textbf{Weight}                                                  & \textbf{Height}                                                    \\
    & \textbf{$[$F/M$]$} &     & \textbf{$[$kg$]$} & \textbf{$[$cm$]$} & \textbf{$[$L/R$]$}  & \textbf{$[$F/M$]$} & $\leq$ 40 / \textgreater 40 & $\leq$ 70 / \textgreater 70 & $\leq$ 170 / \textgreater 170 \\\hline \hline
7   & M         &  23    &  65        &     177     &       R     & 1         &                     0                                    &      0                                                   &       1                                                    \\
8   & F         &   51  &       68   &  168        &        R    & 0         &  1                                                       &                            0                             &            0                                               \\
9   & M         &   35  &  100        & 172         &       R     & 1         &     0                                                    &      1                                                   &  1                                                         \\
10  & M         &  49   &   97       &  181        &        R    & 1         &   1                                                      &             1                                            &    1                                                       \\
11  & F         &  47   &   66       &      175    &    R        & 0         &                    1                                     &         0                                                &    1                                                       \\
12  & F         &   23  &     48     &      163    &      R      & 0         &  0                                                       &        0                                                 &        0                                                   \\
13  & F         &   25  &    54      &  163        &       R     & 0         &   0                                                      &            0                                             &      0                                                     \\
14  & M         &   54  &     90    &      177     &    R        & 1         &    1                                                     &      1                                                   &      1                                                    
\end{tabular}
}
\label{table:attributes}
\end{table}
\begin{figure}[ht]
  \resizebox{1\textwidth}{!}{ 
  \centering
  \begin{tikzpicture} [x=1.0cm,y=0.9cm][scale=0.6]

        \tikzstyle{node1}=[text=black, font=\small \bfseries];
        \tikzstyle{node2}=[text=black, font=\small \bfseries];
        \tikzstyle{node3}=[text=black, font=\small];
        \tikzstyle{node4}=[text=black, font=\footnotesize];
        \tikzstyle{arrow1} = [line width=0.05]
        \tikzstyle{circle1}=[circle,draw=black, minimum size=0.1cm, line width=0.2mm, inner sep=0pt]
        \tikzstyle{circle2}=[circle,draw=black, minimum size=0.05cm, line width=0.1mm, inner sep=0pt, fill=black]
        
        \node [node4] at (0.7,1.7){IMU 1};
        
        \node (label) at (0.85, 0.6)[draw=white, line width=0.0]{
                \includegraphics[width= 0.08\textwidth]{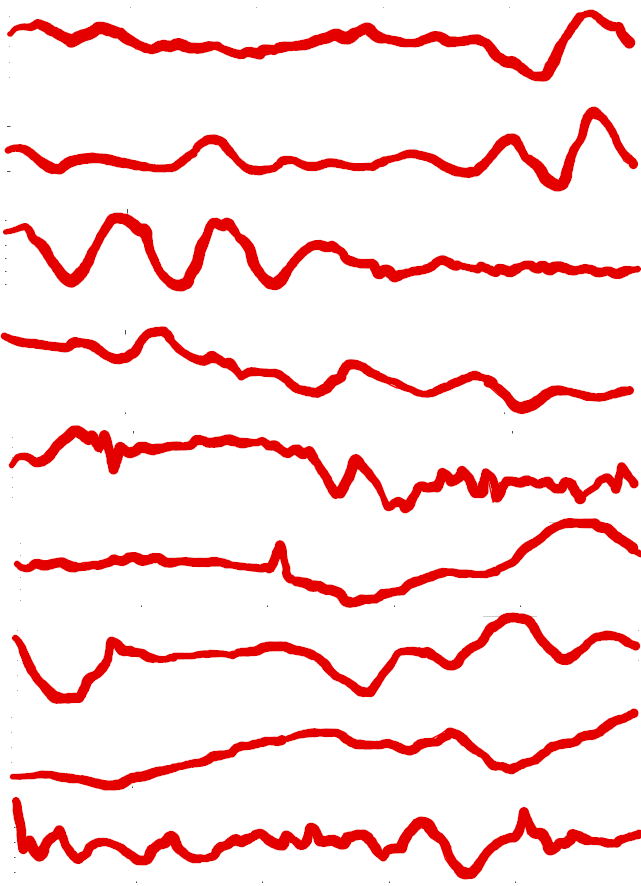}
              }; 
              
        \draw [line width=0.05mm, opacity=0.4] (0.4,0.0 - 0.25) rectangle +(0.9,1.7);
        
        \node [node4] at (2.5-0.5,1.9-0.3){$C=64$};
        
        \draw [line width=0.05mm,fill=lightgray, opacity=0.3](3.0-0.2-0.3,1.5-0.2-0.1)--(2.1-0.3,1.5-0.2-0.1)--(1.8-0.3,1.2-0.2-0.1)--(2.8-0.2-0.3,1.2-0.2-0.1)--(3-0.2-0.3,1.5-0.2-0.1)--(3-0.2-0.3,0.2-0.1)--(2.8-0.2-0.3,-0.1-0.1)--(1.8-0.3,-0.1-0.1)--(1.8-0.3,1.2-0.2-0.1)--(2.8-0.2-0.3,1.2-0.2-0.1)--cycle;
        \draw  [arrow1](3.0-0.2-0.2-0.3,0.2-0.3-0.1) edge (3.0-0.2-0.2-0.3,1.2-0.2-0.1);
        
        \node [node4] at (3.8-0.6,1.9-0.3){$C=64$};
        
        \draw [line width=0.05mm,fill=lightgray, opacity=0.3](4.0+0.3-0.2-0.4,1.5-0.2-0.1)--(3.1+0.3-0.4,1.5-0.2-0.1)--(2.8+0.3-0.4,1.2-0.2-0.1)--(3.8+0.3-0.2-0.4,1.2-0.2-0.1)--(4+0.3-0.2-0.4,1.5-0.2-0.1)--(4+0.3-0.2-0.4,0.2-0.1)--(3.8+0.3-0.2-0.4,-0.1-0.1)--(2.8+0.3-0.4,-0.1-0.1)--(2.8+0.3-0.4,1.2-0.2-0.1)--(3.8+0.3-0.2-0.4,1.2-0.2-0.1)--cycle;
        \draw  [arrow1](4.3-0.2-0.2-0.4,0.2-0.3-0.1) edge (4.3-0.2-0.2-0.4,1.2-0.2-0.1);

        \node [node4] at (5.1-0.7,1.9-0.3){$C=64$};
        
        \draw [line width=0.05mm,fill=lightgray, opacity=0.3](5.0+0.6-0.2-0.5,1.5-0.2-0.1)--(4.1+0.6-0.5,1.5-0.2-0.1)--(3.8+0.6-0.5,1.2-0.2-0.1)--(4.8+0.6-0.2-0.5,1.2-0.2-0.1)--(5+0.6-0.2-0.5,1.5-0.2-0.1)--(5+0.6-0.2-0.5,0.2-0.1)--(4.8+0.6-0.2-0.5,-0.1-0.1)--(3.8+0.6-0.5,-0.1-0.1)--(3.8+0.6-0.5,1.2-0.2-0.1)--(4.8+0.6-0.2-0.5,1.2-0.2-0.1)--cycle;
        \draw  [arrow1](5.4-0.2-0.5,0.2-0.3-0.1) edge (5.4-0.2-0.5,1.2-0.2-0.1);
        
        \node [node4] at (6.4-0.8,1.9-0.3){$C=64$};
        
        \draw [line width=0.05mm,fill=lightgray, opacity=0.3]((6.0+0.9-0.2-0.6,1.5-0.2-0.1)--(5.1+0.9-0.6,1.5-0.2-0.1)--(4.8+0.9-0.6,1.2-0.2-0.1)--(5.8+0.9-0.2-0.6,1.2-0.2-0.1)--(6+0.9-0.2-0.6,1.5-0.2-0.1)--(6+0.9-0.2-0.6,0.2-0.1)--(5.8+0.9-0.2-0.6,-0.1-0.1)--(4.8+0.9-0.6,-0.1-0.1)--(4.8+0.9-0.6,1.2-0.2-0.1)--(5.8+0.9-0.2-0.6,1.2-0.2-0.1)--cycle;
        \draw  [arrow1](6.7-0.2-0.6,0.2-0.3-0.1) edge (6.7-0.2-0.6,1.2-0.2-0.1);

        \node [node4] at (7.6-0.75,0.0+1.7-0.2){$C=256$};
        \draw [fill=lightgray, line width=0.05mm, opacity=0.4] (7.1-0.8,0.0-0.2) rectangle +(0.15,1.5);

        \node [circle2] (c5) at (0.8,1.0-1.5) {};
        \node [circle2] (c6) at (0.8,0.8-1.5) {};
        \node [circle2] (c7) at (0.8,0.6-1.5) {};
        
        \node [circle2] (c5) at (4.1-0.2,1.0-1.6) {};
        \node [circle2] (c6) at (4.1-0.2,0.8-1.6) {};
        \node [circle2] (c7) at (4.1-0.2,0.6-1.6) {};
       
        \node [node4] at (0.7,0.9-2.1){IMU m}; 
        
        \node (label) at (0.85, 0.9 - 3.2)[draw=white, line width=0.0]{ 
                \includegraphics[width= 0.08\textwidth]{Figures/seq_conv.png}
              }; 
              
        \draw [line width=0.05mm, opacity=0.4] (0.4,0.0 - 3.13) rectangle +(0.9,1.7);
        
        \draw [line width=0.05mm,fill=lightgray, opacity=0.3](3.0-0.2-0.3,1.5-3-0.2+0.1)--(2.1-0.3,1.5-3-0.2+0.1)--(1.8-0.3,1.2-3-0.2+0.1)--(2.8-0.2-0.3,1.2-3-0.2+0.1)--(3-0.2-0.3,1.5-3-0.2+0.1)--(3-0.2-0.3,0.2-3+0.1)--(2.8-0.2-0.3,-0.1-3+0.1)--(1.8-0.3,-0.1-3+0.1)--(1.8-0.3,1.2-3-0.2+0.1)--(2.8-0.2-0.3,1.2-3-0.2+0.1)--cycle;
        \draw  [arrow1](3.0-0.2-0.2-0.3,0.2-0.3-3+0.1) edge (3.0-0.2-0.2-0.3,1.2-3-0.2+0.1);

        \draw [line width=0.05mm,fill=lightgray, opacity=0.3](5.0+0.6-0.2-0.5,1.5-3-0.2+0.1)--(4.1+0.6-0.5,1.5-3-0.2+0.1)--(3.8+0.6-0.5,1.2-3-0.2+0.1)--(4.8+0.6-0.2-0.5,1.2-3-0.2+0.1)--(5+0.6-0.2-0.5,1.5-3-0.2+0.1)--(5+0.6-0.2-0.5,0.2-3+0.1)--(4.8+0.6-0.2-0.5,-0.1-3+0.1)--(3.8+0.6-0.5,-0.1-3+0.1)--(3.8+0.6-0.5,1.2-3-0.2+0.1)--(4.8+0.6-0.2-0.5,1.2-3-0.2+0.1)--cycle;
        \draw  [arrow1](5.4-0.2-0.5,0.2-0.3-3+0.1) edge (5.4-0.2-0.5,1.2-3-0.2+0.1);

        \draw [line width=0.05mm,fill=lightgray, opacity=0.3](4.0+0.3-0.2-0.4,1.5-3-0.2+0.1)--(3.1+0.3-0.4,1.5-3-0.2+0.1)--(2.8+0.3-0.4,1.2-3-0.2+0.1)--(3.8+0.3-0.2-0.4,1.2-3-0.2+0.1)--(4+0.3-0.2-0.4,1.5-3-0.2+0.1)--(4+0.3-0.2-0.4,0.2-3+0.1)--(3.8+0.3-0.2-0.4,-0.1-3+0.1)--(2.8+0.3-0.4,-0.1-3+0.1)--(2.8+0.3-0.4,1.2-3-0.2+0.1)--(3.8+0.3-0.2-0.4,1.2-3-0.2+0.1)--cycle;
        \draw  [arrow1](4.3-0.2-0.2-0.4,0.2-0.3-3+0.1) edge (4.3-0.2-0.2-0.4,1.2-3-0.2+0.1);

        \draw [line width=0.05mm,fill=lightgray, opacity=0.3](6.0+0.9-0.2-0.6,1.5-3-0.2+0.1)--(5.1+0.9-0.6,1.5-3-0.2+0.1)--(4.8+0.9-0.6,1.2-3-0.2+0.1)--(5.8+0.9-0.2-0.6,1.2-3-0.2+0.1)--(6+0.9-0.2-0.6,1.5-3-0.2+0.1)--(6+0.9-0.2-0.6,0.2-3+0.1)--(5.8+0.9-0.2-0.6,-0.1-3+0.1)--(4.8+0.9-0.6,-0.1-3+0.1)--(4.8+0.9-0.6,1.2-3-0.2+0.1)--(5.8+0.9-0.2-0.6,1.2-3-0.2+0.1)--cycle;
        \draw  [arrow1](6.7-0.2-0.6,0.2-0.3-3+0.1) edge (6.7-0.2-0.6,1.2-3-0.2+0.1);
        
        \draw [fill=lightgray, line width=0.05mm, opacity=0.4] (7.1-0.8,0.0- 3.2+0.13) rectangle +(0.15,1.5);
        
        \draw [line width=0.05mm, fill=darkgray, opacity=0.1](7.5-0.2-0.85,1.5-0.2)--(7.5-0.2-0.85,-0.3-2.95+0.2)--(7.5-0.2+0.5-0.85,-0.2-1.3)--(7.5-0.2+0.5-0.85,1.6-1.7)--cycle;
        
        \node [node4] at (7.2-0.8,0.0-0.8){concat.};
        
        \draw [fill=darkgray, line width=0.05mm, opacity=0.4] (7.8-0.9,0.0- 1.7) rectangle +(0.2,1.8);
        
        \node [node4] at (8.3-0.8,0.0+0.3){$C=256$};
        \draw [fill=darkgray, line width=0.05mm, opacity=0.4] (8.6-1,0.2- 1.7) rectangle +(0.2,1.4);
        
        \node [node4] at (9.2-1,1.4){Softmax };
        \node [node4] at (9.3-1,1.3-0.2){Identification};
        \draw [fill=purple, line width=0.05mm, opacity=0.4] (9-0.9,0.2- 0.7) rectangle +(0.2,1.4);
        
        \node [node4] at (9.2-0.9,0.0-2.7){Sigmoid};
        \node [node4] at (9.4-0.9,0.0-3){Attribute Representation};
        \draw [fill=blue, line width=0.05mm, opacity=0.4] (9-0.9,0.2- 2.6) rectangle +(0.2,1.4);
        
         \draw [line width=0.05mm, fill=darkgray, opacity=0.1](10-0.9,1.8-1.1)--(10-0.9,-0.3-2.4)--(9.2-0.9,-1-1.4)--(9.2-0.9,1-2)--cycle;
         
         \node (axis) at (11.8-0.7, 0.0 -0.9)[draw=white, line width=0.0]{
               \tikzset{every picture/.style={line width=0.75pt}} 

\begin{tikzpicture}[x=0.75pt,y=0.75pt,yscale=-0.5,xscale=0.5]

\draw  (109,192.4) -- (310,192.4)(132.7,23.2) -- (132.7,211.2) (305,187.4) -- (310,192.4) -- (305,197.4) (127.7,30.2) -- (132.7,23.2) -- (137.7,30.2)  ;
\draw    (132.7,192.4) -- (58.39,269.76) ;
\draw [shift={(57,271.2)}, rotate = 313.85] [color={rgb, 255:red, 0; green, 0; blue, 0 }  ][line width=0.75]    (10.93,-3.29) .. controls (6.95,-1.4) and (3.31,-0.3) .. (0,0) .. controls (3.31,0.3) and (6.95,1.4) .. (10.93,3.29)   ;
\draw  [dash pattern={on 0.84pt off 2.51pt}] (132.7,88.2) -- (243,88.2) -- (243,192.4) -- (132.7,192.4) -- cycle ;
\draw  [fill={rgb, 255:red, 208; green, 2; blue, 27 }  ,fill opacity=1 ] (243,78.3) -- (244.8,85.48) -- (251.56,85.14) -- (245.91,89.24) -- (248.29,96.21) -- (243,91.57) -- (237.71,96.21) -- (240.09,89.24) -- (234.44,85.14) -- (241.2,85.48) -- cycle ;
\draw  [dash pattern={on 0.84pt off 2.51pt}] (132.87,88.36) -- (205.82,159.05) -- (205.65,263.09) -- (132.7,192.4) -- cycle ;
\draw  [fill={rgb, 255:red, 74; green, 144; blue, 226 }  ,fill opacity=1 ] (207.61,151.88) -- (209.41,159.05) -- (216.17,158.72) -- (210.52,162.82) -- (212.9,169.79) -- (207.61,165.14) -- (202.32,169.79) -- (204.7,162.82) -- (199.05,158.72) -- (205.82,159.05) -- cycle ;
\draw  [dash pattern={on 0.84pt off 2.51pt}]  (243,192.4) -- (205.65,263.09) ;
\draw  [dash pattern={on 0.84pt off 2.51pt}]  (78,249.2) -- (205.65,263.09) ;

\tikzstyle{node4}=[text=black, font=\footnotesize];


\node [node4] at (138,20) [anchor=north west][inner sep=0.3pt]    {$Att_{1}$};
\node [node4] at (270,195) [anchor=north west][inner sep=0.75pt]    {$Att_{2}$};
\node [node4] at (53,265) [anchor=north west][inner sep=0.75pt]    {$Att_{3}$};
\node [node4] at (235,196) [anchor=north west][inner sep=0.75pt]    {$1$};
\node [node4] at (243,65) [anchor=north west][inner sep=0.75pt]    {$Ind_{x}$};
\node [node4] at (200,133) [anchor=north west][inner sep=0.75pt]    {$Ind_{y}$};
\node [node4] at (163,165) [anchor=north west][inner sep=0.75pt]    {$Ind_{z}$};
\node [node4] at (65,228) [anchor=north west][inner sep=0.75pt]    {$1$};
\node [node4] at (116,76) [anchor=north west][inner sep=0.75pt]    {$1$};

\end{tikzpicture} 
              };
    \end{tikzpicture}
  }
  \caption{The tCNN-IMU architecture contains $m$ parallel temporal-convolutional blocks, one per limb. The outputs of the blocks are concatenated and forwarded to a fully connected or 2-LSTM layers according to the network type. Person identification uses softmax in the output layer. Attribute representation uses sigmoid function.}
  \label{fig:networks}
\end{figure}

This work investigates person identification from multi-channel time series recordings of humans performing various activities. Though gait movements are considered to be biometric, body movements performed during activities of daily living have not been recognised to be unique. However, classifiers can identify individuals when performing activities following \cite{elkaderWearableSensorsRecognizing2018}. Here, we attempt to provide a standardised and reproducible set of experiments. Firstly, we explore the impact of channel-normalisation. It is expected that positional bias of the sensors may help improve person identification. Next, we explore the possibility of generalisation of identity over activities. The expectation is that similar activities generalise based on the individual's motion signature. As a result, we employ DNNs on OBD recordings of human activities to attempt at person and soft-biometrics identification. Soft-biometrics identification will be experimented with by modelling soft-biometrics as an attribute representation. Finally, we perform LRP on select dataset to explore the feasibility of XAI to verify person identification.

Tab.~\ref{table:attributes} and \ref{fig:networks} briefs the method. Firstly, the OBD data is grouped based on the limb that the OBD is placed on. For example, if the limb \textit{left leg} has two OBDs placed on the ankle and knee respectively, the OBDs will be grouped together. Thus, each limb of the human body with an OBD will be allotted a block of four convolutional layers. This allows extraction of local features. Next the layers are concatenated and followed by two fully connected layers which extract the global features. The final layer is derived based on the required process; person identification using a softmax layer or soft-biometrics using a sigmoid layer. For attribute representation, a person is represented by either a one-hot encoding or a set of soft-biometrics. The attribute representation is created based on the recording protocol. 
\subsection{Networks}

We use the \textit{tCNN-IMU} network \cite{grzeszickDeepNeuralNetwork2017, moyaruedaConvolutionalNeuralNetworks2018, ruedaLearningAttributeRepresentation2018a}. Here, the late fusion approach is applied using an MLP (\textit{tCNN-IMU}$_{MLP}$) or a two-layered LSTM (\textit{tCNN-IMU}$_{LSTM}$), with a softmax or sigmoid classifier. OBDs from each human limb is allotted a branch of four convolutional layers. The convolutional blocks extract descriptive local features from the input OBD data, while the subsequent layers assimilate the global view of the extracted features. Prior to concatenation, the outputs of each convolutional blocks are processed by a fully connected layer or LSTM layer depending on the network type. Concatenation is followed by a two-layered fully connected MLP and a classifier layer for \textit{tCNN-IMU}$_{MLP}$. In the case of \textit{tCNN-IMU}$_{LSTM}$, the concatenation is followed by two LSTM layers and a classifier layer. 
        
The networks use a softmax classifier for person identification, whereas a sigmoid layer for soft-biometrics identification \cite{ruedaLearningAttributeRepresentation2018a}. Soft-biometrics of individuals describe or categorise an individual or a group of individuals \cite{shahidStudyHumanGait2012}, e.g., Gender Identity, Age, Weight and Height. Attribute representations is a method of describing the data semantically \cite{ruedaLearningAttributeRepresentation2018a}. An attribute vector $a$ represents a set of soft-biometrics. A similar combination of soft-biometrics could represent different persons with similar features. The Nearest Neighbour Approach (NNA) is used for soft-biometrics-based identification. The NNA calculates the distance between a prediction attribute vector $a$ and an attribute representation $A$, with all the different combinations of soft-biometrics. The person identity is assigned to the one related with the least distance from $A$. NNA is performed by computing a certain similarity between a $A$ and the vector $a$ from the network; typically, the \emph{cosine} similarity \cite{sudholtAttributeCNNsWord2018} and the \emph{Probabilistic Retrieval Model (PRM)} similarity \cite{rusakovProbabilisticRetrievalModel2018}.

\subsubsection*{Training Procedure}
        
The weights of the network are initialised using the orthogonal initialisation method. The Cross-Entropy Loss function is utilised to calculate person classification loss. In the case of attribute representation, the $BCE_{loss}$ is used. The Root Mean Square Propagation (RMSProp) optimisation is used with a momentum of $0.9$ and weight decay of $5\times10^{-4}$. Gaussian noise with mean $\mu = 0$ and SD $\sigma = 0.01$ is added to the sensor measurements to simulate sensor inaccuracies \cite{moyaruedaConvolutionalNeuralNetworks2018, grzeszickDeepNeuralNetwork2017}. Dropout of probability $p=0.5$ was applied on the MLP and LSTM layers of the networks and early-stopping to avoid over-fitting.
            
\section{Experiments and Results}
Person identification was performed on four HAR datasets, namely LARa \cite{niemannLARaCreatingDataset2020}, OPPORTUNITY (OPP) \cite{chavarriagaOpportunityChallengeBenchmark2013}, PAMAP2 \cite{reissIntroducingNewBenchmarked2012}, and Order Picking (OrP) \cite{grzeszickDeepNeuralNetwork2017} using the \textit{tCNN-IMU}$_{MLP}$ and \textit{tCNN-IMU}$_{LSTM}$. The datasets were selected based on their public availability and prominence in HAR research, thus facilitating comparable and competitive research. Additionally, soft-biometrics identification was performed on the LARa and PAMAP2 datasets using \textit{tCNN-IMU}$_{MLP}$; this as its recording protocol is available. To create the training, validation and test set, the recordings of the individuals are split as per $64-18-18\%$. The split percentages were decided based on the number of recordings available for each individual, while ensuring closeness to standard split percentages. 

\subsection{Datasets}
    
The datasets of interest are different from each other with respect to the experimental set-up, recording protocol and activities. Table.~\ref{table:datasetcompare} tabulates the differences between the datasets based on the number of subjects, IMUs, channels and activities present in the dataset. Furthermore, the table presents the differences in IMU placement, recording location and activity types. It is to be noted that OrP was recorded from two different warehouses, where one subject was common to both warehouses. Furthermore, the dataset does not have a recording protocol. Unlike laboratory-made LARa and OPP datasets, PAMAP2 requires pre-processing to overcome data loss.
    
\begin{table}[h]
\centering
\caption{Comparison of the features of selected datasets based on number of subjects, IMUs, channels, position of OBDs, and recording environment.}
\resizebox{1.0\columnwidth}{!}{
\begin{tabular}{|c|c|c|c|c|c|c|}
    \hline
    \textbf{Dataset}           & \textbf{MoCap} & \textbf{Mbientlab} &  \begin{tabular}[c]{@{}c@{}}\textbf{Motion}\\ \textbf{Miners}\end{tabular} & \textbf{OPP}  & \textbf{PAMAP2} & \textbf{OrP} \\ \hline
    \textbf{Subject No:}   & 14                   & 8                      & 8                          & 4                     & 9               & 6                       \\ \hline
    \textbf{IMU No:}       & -                  & 5                       & 3                           & 7                     & 3              & 3                        \\ \hline
    \textbf{Channel No:}   & 126                & 30                     &  27                          & 113                   & 40            & 27                        \\ \hline
    \begin{tabular}[c]{@{}c@{}}\textbf{Additional} \\ \textbf{Sensors:}\end{tabular} & -                  & -                     & -                           & Accelerometers          & \begin{tabular}[c]{@{}c@{}}Heart \\ Monitor\end{tabular}  & -                         \\ \hline
    \textbf{Activities No:} & 7                 & 7                      & 7                         & \begin{tabular}[c]{@{}c@{}}Locomotion:4\\ Gesture:17\end{tabular}  & 12   & 7  \\ \hline
    \begin{tabular}[c]{@{}c@{}}\textbf{IMU} \\ \textbf{Placement}\end{tabular}     & \begin{tabular}[c]{@{}c@{}}Chest, \\ wrists, legs\end{tabular} & \begin{tabular}[c]{@{}c@{}}Chest, \\ wrists, legs\end{tabular}  & \begin{tabular}[c]{@{}c@{}}Hip, \\ wrists \end{tabular} & \begin{tabular}[c]{@{}c@{}}Chest, wrists,\\ arms, legs\end{tabular} & \begin{tabular}[c]{@{}c@{}}Chest,\\ right wrist\\ right ankle\end{tabular} & \begin{tabular}[c]{@{}c@{}}Chest, \\  wrists\end{tabular} \\ \hline
    \textbf{Location}          & Lab              & Lab                    & Lab                        & Lab - Kitchen             & Outdoor       & Warehouses                    \\ \hline
    \textbf{Act. Type}    & Logistics       & Logistics                 & Logistics                 & ADL                       & ADL           & Logistics                     \\ \hline
    \textbf{Sampl. Rate:} & 200 Hz        & 100 Hz        & 100Hz             & 30 Hz             & 100 Hz            & 100Hz \\ \hline
\end{tabular}
}
\label{table:datasetcompare}
\end{table}

\subsubsection*{Pre-processing}
        
Sliding window size equivalent to step frequency or duration of $1$sec. provides better results according to \cite{goharPersonReIdentificationUsing2020}. Consequently, a sliding window size corresponding to the sampling rate was considered for the datasets. LARa-Mbientlab, LARa-MotionMiners and PAMAP2 have a sliding window size of $100$ frames and a stride size of $12$ frames. OPP was experimented with two sliding window sizes, $24$ and $100$ frames ($\sim 4$sec.) and a constant stride size of $12$ frames. OrP has a fixed window size of $100$ frames and stride $1$. A zero-mean and unit-SD channel-normalisation is carried out as part of the training procedure from \cite{ruedaLearningAttributeRepresentation2018a}, as networks' filters are shared among the channels, independent of their magnitude.

\subsection{Person Identification on LARa}
    
Tab.~\ref{table:compareoi} shows the performance of person identification in terms of Accuracy (Acc) [$\%$] and weighted $F1$ (wF1)[$\%$] on the LARa-Mbientlab and LARa-MotionMiners using the two architectures, \textit{tCNN-IMU}$_{MLP}$ and \textit{tCNN-IMU}$_{LSTM}$ with three learning rates, $Lr$=$[10^{-4},10^{-5},10^{-6}]$, three batch sizes mB size=$[50,100,200]$ and $10$ epochs. All experimental results are an average of five trial runs \textbf{(x5)} and are presented as percentages. The experiment emphasises the person identification performance on different sensor sets recording similar activities. Both the networks are trained on LARa-Mbientlab and LARa-MotionMiners under similar training conditions. Person identification was found to be feasible for LARa with both five and three OBDs. \textit{tCNN-IMU}$_{MLP}$ outperforms \textit{tCNN-IMU}$_{LSTM}$. Interestingly, a smaller batch size performs better. Though LARa-MotionMiners have three sensors on the upper body, the accuracy of person identification is comparable to the LARa-Mbientlab with five sensor points. Furthermore, Tab.~\ref{table:compareoi} presents the performance for LARa-Mocap, considering the human poses as multi-channel time series, using the \textit{tCNN-IMU}$_{MLP}$. Person identification performance decreases when using OBDs in comparison to human poses. However, in the best scenario using the \textit{tCNN-IMU}$_{MLP}$ for both OBDs sets, the performance decreases only by $\sim 3\%$; this, as OBDs and human poses are physical related quantities.

\begin{table}[h]
    \centering
    \caption{Person Identification in terms of the average Acc[$\%$] and wF1[$\%$] from five runs \textbf{(x5)} on the LARa-Mbientlab and LARa-MotionMiners using the \textit{tCNN-IMU}$_{MLP}$ and \textit{tCNN-IMU}$_{LSTM}$ networks and different batches at $Lr$=$10^{-4}$ and epoch $10$.  Here, *$_{MLP}$ and *$_{LSTM}$ represent tCNN-IMU$_{MLP}$ and tCNN-IMU$_{LSTM}$ respectively.}
    \resizebox{\columnwidth}{!}{
        \begin{tabular}{c|c| cc |cc}
            \multirow{2}{*}{\textbf{Network}} & \multirow{2}{*}{\textbf{mB}} & \multicolumn{2}{c}{\textbf{LARa-Mbientlab}} & \multicolumn{2}{c}{\textbf{LARa-MotionMiners}} \\
    
             &  & \textbf{Avg Acc (x5)}        & \textbf{Avg wF1 (x5)} & \textbf{Avg Acc (x5)}        & \textbf{Avg wF1 (x5)}     \\ 
            \hline
            \hline
            \multirow{3}{*}{*$_{MLP}$}  & 50       & \textbf{93.96 $\pm$ 0.03} & \textbf{93.84 $\pm$ 0.03} & \textbf{93.32 $\pm$ 0.21} & \textbf{93.32 $\pm$ 0.22} \\ 
            & 100   & 92.17 $\pm$ 0.003 & 92.11 $\pm$ 0.39 & 91.36 $\pm$ 1.67 & 91.67 $\pm$ 1.67  \\ 
            & 200               & 90.15 $\pm$ 0.20          & 90.01 $\pm$ 0.21 & 88.74 $\pm$ 0.92          & 88.29 $\pm$ 0.92           \\
            \hline
            \multirow{3}{*}{*$_{LSTM}$} & 50                & \textbf{90.55 $\pm$ 0.57}   & \textbf{90.41 $\pm$ 0.61}   & 86.19 $\pm$ 0.76  & 86.21 $\pm$ 0.76    \\ 
            & 100               & 89.43 $\pm$ 0.96   & 89.22 $\pm$ 1.01  & \textbf{86.43 $\pm$ 0.53}   & \textbf{86.41 $\pm$ 0.52}    \\ 
            & 200               & 86.48 $\pm$ 0.86   & 86.08 $\pm$ 0.88  & 84.12 $\pm$ 2.09   & 84.02 $\pm$ 2.17 \\ 
            
            \multicolumn{4}{c}{\vspace{-2mm}}\\
            
            \multirow{2}{*}{\textbf{Network}} & \multirow{2}{*}{\textbf{mB}} &  \multicolumn{4}{c}{\textbf{LARa-MoCap}} \\
            & &  \multicolumn{2}{c}{\textbf{Avg Acc (x5)}}         &  \multicolumn{2}{c}{\textbf{Avg wF1 (x5)}}      \\ 
            \hline
            \hline
            *$_{MLP}$  & 100   & \multicolumn{2}{c}{96.72 $\pm$ 0.72}  & \multicolumn{2}{c}{ 96.71 $\pm$ 0.72}\\ 
            \hline
    \end{tabular}
    }
    \label{table:compareoi}
\end{table}
        
Tab.~\ref{table:compareoi} shows the results using normalised channels; this, as network process sequences time-wise, sharing its filters channel-wise. Consequently, non- and channel-normalised data are compared. This comparison follows the assumption that non-channel-normalised data might provide better person identification performance due to the sensor placement bias. Contrary to expectation, the channel-normalised data provided better identification performance, as shown in Tab.~\ref{table:type4IMU}. Similar to Tab.~\ref{table:compareoi}, a smaller batch size provides better performance.
        
\begin{table}[ht]
    \centering
    \caption{Comparison of channel-normalised and non-channel-normalised LARa-Mbientlab on \textit{tCNN-IMU}$_{MLP}$. Average Acc[\%] and wF1[\%] of five runs \textbf{(x5)} with $Lr$=$10^{-4}$ and epoch $10$.}
    \resizebox{\columnwidth}{!}{
            \begin{tabular}{c|c c |c c}
                
                \multirow{2}{*}{\textbf{mB}}& \multicolumn{2}{c|}{\textbf{Non-Norm}} & \multicolumn{2}{c}{\textbf{Norm}} \\
                 & \textbf{Avg Acc (x5)}  & \textbf{Avg wF1(x5)} & \textbf{Avg Acc (x5)}        & \textbf{Avg wF1(x5)}     \\ 
                \hline
                \hline
                50       &  90.06 $\pm$ 0.65 & 89.97 $\pm$ 0.69 & 93.96 $\pm$ 0.03 & 93.84 $\pm$ 0.03\\ 
                100     & 88.59 $\pm$ 0.65 & 88.41 $\pm$ 0.73  &  92.17 $\pm$ 0.003 & 92.11 $\pm$ 0.39 \\ 
                200               & 84.51 $\pm$ 0.73          & 84.06 $\pm$ 0.89  & 90.15 $\pm$ 0.20          & 90.01 $\pm$ 0.21          \\ 
                \hline
            \end{tabular}
            }
    \label{table:type4IMU}
\end{table}
        
\begin{figure}[h]
    \centering
    \begin{tabular}{cc}
        \includegraphics[height=3.5cm]{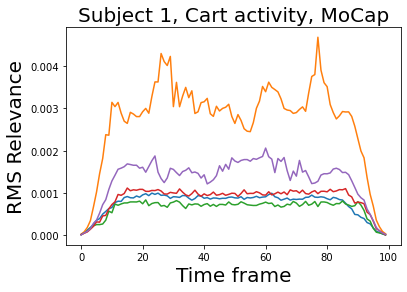} & \hspace{-0.3cm}\includegraphics[height=3.5cm]{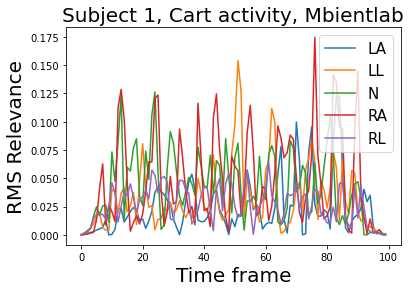} \\
        \includegraphics[height=3.5cm]{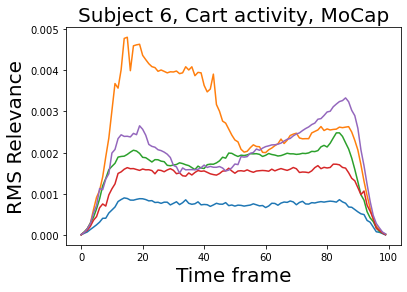} & \hspace{-0.3cm}\includegraphics[height=3.5cm]{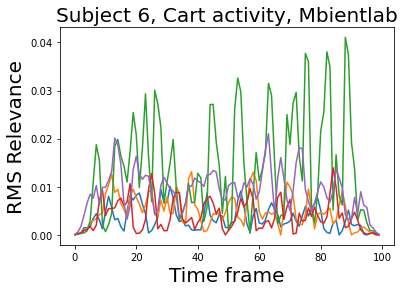} \\
    \end{tabular}
    \caption{LRP on LARa dataset with a \textit{tCNN-IMU}$_{MLP}$ of one branch for all channels. The plots show the RMS of positive relevance of the sensor channels of Subject $1$ and $6$ performing the activity "Cart".}
    \label{fig:cartcomp2}
\end{figure}

As a preliminary work towards analysing the features that contribute to person identity within HAR data, $\epsilon$-LRP \cite{horstExplainingUniqueNature2019, montavonExplainingNonlinearClassification2017, montavonLayerWiseRelevancePropagation2019} was attempted on the OBDs (LARa-Mbientlab) and human poses (LARa-Mocap) per limb with a \textit{tCNN-IMU}$_{MLP}$ of one branch for all channels. The $\epsilon$ value was fixed at $10^{-9}$. Fig.~\ref{fig:cartcomp2} shows the root mean square (RMS) of the positive relevances of the limbs (LARa-Mbientlab, LARa-Mocap) of subjects $1$ and $6$, as mentioned in LARa-Mbientlab protocol, performing the activity "Cart". In this activity, the subjects walk, transporting a Cart with both or one hand. The limbs are represented as left leg (LL), left arm (LA), Neck/Torso (N), right arm (RA) and right leg (RL). For LARa-Mocap, the human limbs contributing to the person identification are related to the "Cart" activity; that is the right and left legs. For LARa-Mbientlab, the Neck/Torso area is relevant for the correct identification. These plots are only a single example of $1$sec. window of an activity. However, the patterns and variations point that accumulative analysis using tools as mentioned in \cite{andersSoftwareDatasetwideXAI2021} might help develop methods to either mask or delete the identity data or enhance HAR datasets. 
        
\subsection{Person Identification on PAMAP2}
    
While initially experimenting with PAMAP2, each recording of the individuals was split into the train-validation-test (T-V-T) set at $64\%$-$18\%$-$18\%$ without considering the activity being performed in the segmented windows of the recordings. The results on the dataset were similar irrespective of the $Lr$, mB size, window size and stride at around Acc $50\%$. The maximum Acc was $57.2\%$ for $Lr$=$10^{-4}$, mB size=$50$, epoch $50$ and without subject $109$, as in the recording protocol of PAMAP2. 
        
A new training set was created to investigate whether the poor identification performance was related to the distribution of activities throughout the T-V-T set. The recordings per activity were stacked and then split into $[64\%$-$18\%$-$18\%]$ T-V-T sets. Consequently, every activity performed by the individual in the test set would be present in the training, validation and testing set. It was found that the new T-V-T set could perform better and provide better identification performance, as shown in Tab.~\ref{table:pamapid}. The results are an average over five trials \textbf{(x5)} and are presented as percentages. The experiment indicates that the network may not generalise features contributing to an individual's identity over activities that are not present in the training set irrespective of their similarity to the activities. The \textit{tCNN-IMU}$_{LSTM}$ shows a better and lesser variable performance than the \textit{tCNN-IMU}$_{MLP}$.
        
\begin{table}[h]
    \centering
    \caption{Person Identification in terms of the average Acc[$\%$] and wF1[$\%$] from five runs \textbf{(x5)} on the PAMAP2 and OrP using the \textit{tCNN-IMU}$_{MLP}$ and \textit{tCNN-IMU}$_{LSTM}$ at $Lr=10^{-4}$ for different batches. PAMAP2 runs for 10 epochs and OrP for 4 epochs. The original split with "*" was for $50$ epochs. Here, *$_{MLP}$ and *$_{LSTM}$ represent tCNN-IMU$_{MLP}$ and tCNN-IMU$_{LSTM}$ respectively.}
    \resizebox{\columnwidth}{!}{
        \begin{tabular}{c|c|cc | cc}
           
            & &\multicolumn{2}{c}{\textbf{PAMAP2}} & \multicolumn{2}{c}{\textbf{Order Picking}} \\
           
            \textbf{Network} & \textbf{mB} & \textbf{Avg Acc (x5)}        & \textbf{Avg wF1 (x5)}  & \textbf{Avg Acc (x5)}        & \textbf{Avg wF1 (x5)}      \\ 
            \hline
            O. Split*  & 50       & 54.93 $\pm$ 4.29 & 53.84 $\pm$ 5.24 & & \\
            \hline
            \hline
            \multirow{3}{*}{*$_{MLP}$}  & 50       & \textbf{90.35 $\pm$ 0.61} & \textbf{90.36 $\pm$ 0.61} & 52.79 $\pm$0.98 & 51.69 $\pm$0.92 \\ 
            & 100   & 85.03 $\pm$ 0.43 & 84.98 $\pm$ 0.43 & 52.80 $\pm$0.41 & 51.71 $\pm$ 0.34  \\ 
            & 200               & 75.74 $\pm$ 1.45          & 75.63 $\pm$ 1.51 & \textbf{53.27 $\pm$ 2.11} & \textbf{51.72 $\pm$ 1.58}         \\
            \hline
            \multirow{3}{*}{*$_{LSTM}$} & 50                & \textbf{91.65 $\pm$ 0.25}   & \textbf{91.64 $\pm$ 0.25}  & 49.30 $\pm$2.83 & 47.42$\pm$3.39     \\ 
            & 100               & 90.31 $\pm$ 0.07   & 90.30 $\pm$ 0.08    & 50.09 $\pm$ 1.11 & 48.56 $\pm$ 1.70    \\ 
            & 200               & 87.88 $\pm$ 0.89   & 87.87 $\pm$ 0.90  & \textbf{50.17 $\pm$ 1.81} & \textbf{49.16 $\pm$ 1.85}\\ 
            \hline
        \end{tabular}
    }
    \label{table:pamapid}
\end{table}
     
\subsection{Person Identification on OPPORTUNITY}
    
A \textit{tCNN-IMU}$_{MLP}$ was trained on OPP with epochs = $\{5, 10\}$, mB sizes=$\{25, 100\}$, and $Lr$=$\{10^{-4}, 10^{-5}, 10^{-6}\}$. An Acc and wF1 of $99\%$ was achieved for mB sizes=$\{25, 100\}$, of $Lr$=$10^{-4}$ with epochs $5$ and $10$. Similar to LARa-Mbientlab data, it was noticed that reducing the $Lr$ to $10^{-6}$ deteriorated the performance of the networks. However, the Acc remained greater than $90\%$. An average Acc of $96.03\%$ and an average wF1 of $95.84\%$ was achieved five runs \textbf{(x5)} of the experiment. The performance of the \textit{tCNN-IMU}$_{LSTM}$ was similar to the \textit{tCNN-IMU}$_{MLP}$ network.

\subsection{Person Identification on Order Picking}
    
OrP is a small dataset created in two real warehouses for HAR. Irrespective of the $Lr$, mB size and epoch, the Acc failed to improve beyond $55\%$ on both networks. Further, we attempted to find the accuracy on the non-normalised dataset with epoch $15$. This resulted in an accuracy level of $57\%$. Furthermore, we attempted at fine-tuning the dataset using the convolutional layers of the \textit{tCNN-IMU}$_{MLP}$ trained on LARa-Mbientlab. However, no improvements in identification were achieved. The absence of a recording protocol limits the chances of further experimentation and improvements. Tab.~\ref{table:pamapid} presents the person identification results on OrP with the two networks. The training was conducted at $Lr$=$10^{-4}$ and epoch $4$. The values presented are an average of five iterations. 
    
\subsection{Impact of Activities}
        
We measured the proportion of activities when an identity is correctly predicted on the test sets; that is $IOA_{c}^+ = \frac{n_{c}^+}{n_{c}^+ + n_{c}^-}$ for an activity class $c$. Here, $n_{c}^+$ refers to number of windows with correct person identification and $n_{c}^-$ refers to the number of windows with the activity label that were misclassified. 
  
Tab.~\ref{table:iolara} shows the $IOA_{c}^+$ for PAMAP2, LARa-Mbientlab, and OPP. For LARa-Mbientlab, the activities that contain gait cycles performed better than activities with upper body movement. The worst performance was showcased by the "Handling down" activity. For OPP, the windows with minimal body movements obtained a higher correct identification rate, similar to the results of \cite{elkaderWearableSensorsRecognizing2018}. On the other hand, the windows with "Stand" locomotion activity had the least correct window classification. Overall the performance of OPP dataset was good because of the sensor-rich nature of the dataset.

\begin{table}[ht]
        \caption{Impact of activities on PAMAP2, LARa and OPP datasets. The averaged accuracy values are presented as a percentage.}
        \label{table:iolara}
        \centering
        \resizebox{0.7\textwidth}{!}{
        \begin{tabular}{cc cc}
         \multicolumn{2}{c}{\textbf{LARa}} & \multicolumn{2}{c}{\textbf{PAMAP2}} \\
        \textbf{Activity}  & \textbf{$IOA_{c}^+$} & \textbf{Activity}  & \textbf{$IOA_{c}^+$} \\
        Walking       & 87.67 $\pm$ 12.23  & Rope Jump.  & \textbf{83.74 $\pm$ 6.81}  \\ 
        Cart         & \textbf{93.37 $\pm$ 6.37} & Lying  & 74.72 $\pm$ 26.85   \\ 
        Hand. cen  & 83.08 $\pm$ 8.02 & Sitting & 75.58 $\pm$ 13.97  \\ 
        Hand. down & 66.74 $\pm$ 22.58 & Standing & 78.73 $\pm$ 14.46  \\ 
        Hand. up   & 82.187 $\pm$ 13.27 & Walking & \textbf{85.65 $\pm$ 19.15} \\ 
        Standing      & \textbf{89.21 $\pm$ 8.90} & Running & 77.41 $\pm$ 12.76  \\ 
        Synch        & 85.02 $\pm$ 9.48 & Cycling & 68.64 $\pm$ 22.18  \\ 
        \multicolumn{2}{c}{\textbf{OPP}} & Nordic Walk & \textbf{87.23 $\pm$ 13.98}  \\ 
        \textbf{Activity}      & \textbf{$IOA_{c}^+$} & Asc. Stairs & 74.09 $\pm$ 22.46  \\ 
        None   & 98.64 $\pm$ 1.68 & Des. Stairs & 66.64 $\pm$ 24.77  \\ 
        Stand   & 92.42 $\pm$ 3.93 & Vaccuming        & 49.39 $\pm$ 19.84   \\ 
        Walk  & 97.69 $\pm$ 2.83 & Ironing           & 74.91 $\pm$ 22.71  \\ 
        Lie & 98.04 $\pm$ 5.72 & -          & -   \\ 
        Sit   & 98.82 $\pm$ 2.53 & -          & -  \\ 
        \end{tabular}
        }
\end{table}

From the four basic activities in PAMAP2 with cyclic body movements, "Cycling" activity shows a poor classification rate. "Nordic walk" and "Walking" perform better than all other activity classes. "Lying", "Sitting" and "Standing" are classified as postures.  In general, the classification rate of postures is low. This finding negates the conclusion of \cite{elkaderWearableSensorsRecognizing2018}, that activities with little body movement have high identification accuracy. Identical to the findings in \cite{elkaderWearableSensorsRecognizing2018}, the classification accuracy of "Vacuuming" activity was the worst performance. Activities with tools that cause vibrations impact the sensor data, leading to poor classification accuracy. 

One anomaly of the entire dataset is descending stairs activities. Though "Ascending Stairs" has an average performance, "Descending Stairs" has a similar performance to that of "Cycling". It is unclear why two activities that show cyclic body movement showed accuracy rates worse than that of posture activities. 
        
\subsection{Soft-Biometrics Identification}

Taking inspiration from \cite{riazOneSmallStep2015, ruedaLearningAttributeRepresentation2018a, reiningAttributeRepresentationHuman2018}, two sets of attribute representations $[A_1, A_2]$ are created for person identification based on soft-biometrics from the recording protocol of LARa dataset. $A_1 \in \mathbb{B}^4$ contains four binary attributes, which are quantised into two levels. For example, the attribute $A_1^{height}$=$0$ when the height is $\leq170cm$, and $A_1^{height}$=$1$ when the height is \textgreater$170cm$. $A_2 \in \mathbb{B}^{10}$ contains $10$ binary attributes representing the soft-biometrics \footnote{The attribute representation for the two types can be found in \url{https://github.com/nilahnair/Annotation_Tool_LARa/tree/master/From_Human_Pose_to_On_Body_Devices_for_Human_Activity_Recognition/Person_SoftBio_Identification}.}. Each soft-biometric is divided into three levels, e.g, $A_2^{height[\leq170, 170-180, >180]}$. In this case, each level is assigned a binary value, e.g., given an individual of height $163$, $A_2^{height[\leq170]}$=$1$, $A_2^{height[170-180]}$=$0$ and $A_2^{height[>180]}$=$0$. Given the limited number of subjects in the LARa dataset, we selected the attributes and thresholds such that they contain variations that can facilitate learning. For example, the protocol contains information regarding handedness; however, it was not considered an attribute for learning as there were no variations  within the subjects, i.e., all subjects were right-handed. Consequently, applying the same attribute representation on a different dataset, for instance, PAMAP2, would not be feasible, as the ranges of the attributes would be different; thus, resulting in a representation with no variation.

\begin{table}[!htb]
    \centering
    \caption{Leave one out cross-validation (LOOCV) attribute representations using NNA for LARa-Mbientlab, LARa-Motionminers and PAMAP2. M/F refers to Male or Female. L/R refers to handedness left or right. Lr rate= 0.001 and epoch 50. The accuracy values are presented as a percentage.}
    \label{Table: t1cvNN}               
    \resizebox{1\columnwidth}{!}{
    \begin{tabular}{c|ccc|ccc|cc}
              & & $A_1$ & & & $A_2$ & & & \\
               & \textbf{Threshold}     & \textbf{Mbientlab} & \textbf{MotionMiners} & \textbf{Threshold}     & \textbf{Mbientlab} & \textbf{MotionMiners} &  \textbf{Threshold}     &  \textbf{PAMAP2}      \\ 
                \hline
                \hline
                \textbf{Gender}   & M/F  & 44.81 $\pm$ 24.21 & \textbf{46.03 $\pm$ 37.44} & M/F & \textbf{47.19 $\pm$ 24.43} &  44.15 $\pm$ 40.54  & M/F & \textbf{81.11 $\pm$ 32.17}\\
                \hline
                \textbf{Handedness}   & L/R   &   &  & L/R & &   & L/R & \textbf{83.82 $\pm$ 32.19} \\
                \hline
                \multirow{3}{*}{\textbf{Age}}   &  \multirow{3}{*}{$\leq$40$/>$40}   & \multirow{3}{*}{ 35.11 $\pm$ 19.21} & \multirow{3}{*}{ \textbf{64.94 $\pm$ 29.84}}  &$\leq$30  & 44.22 $\pm$ 29.11 & \textbf{45.77 $\pm$ 34.61} & $\leq$25  & 33.37 $\pm$ 15.95 \\ 
                & & &  &30-40 & 66.87 $\pm$ 29.43 &  \textbf{76.08 $\pm$ 42.24} & 25-30 & 44.99 $\pm$ 21.45 \\ 
                & & & & $>$40 & 38.44 $\pm$ 22.49 & \textbf{66.55 $\pm$ 31.94} & $>$30 & \textbf{56.82 $\pm$ 42.11}\\ 
                \hline
                \multirow{3}{*}{\textbf{Weight}}   & \multirow{3}{*}{$\leq$70$/>$70}   & \multirow{3}{*}{43.82 $\pm$ 20.98}  & \multirow{3}{*}{\textbf{58.84 $\pm$ 35.83}} & $\leq$60    & \textbf{46.26 $\pm$ 32.27}  & 30.79 $\pm$ 35.53  & $\leq$70 & 63.58 $\pm$ 34.09 \\ 
                &&& & 60-80 & \textbf{55.18 $\pm$ 43.92} & 48.52 $\pm$ 42.06 & 70-80  & \textbf{66.35 $\pm$ 31.08}\\ 
                & & & & $>$80 & 47.62 $\pm$21.32 & \textbf{59.13 $\pm$ 39.83} & $>$80 & 52.67 $\pm$ 19.63\\
                \hline
                \multirow{3}{*}{\textbf{Height}}   &\multirow{3}{*}{$\leq$170$/>$170}   & \multirow{3}{*}{\textbf{47.77 $\pm$ 30.78}} & \multirow{3}{*}{35.82 $\pm$ 34.17} & $\leq$170  & \textbf{50.89 $\pm$ 30.56} & 33.78 $\pm$ 37.00  & $\leq$175 & 50.10 $\pm$ 28.86 \\ 
                & & & & 170-180 & \textbf{40.83 $\pm$ 24.55} & 31.48 $\pm$ 39.44 & 175-185 & 44.84 $\pm$ 21.75\\ 
                & & & & \textgreater180 & 65.96 $\pm$ 33.17 & \textbf{72.09 $\pm$ 42.25} & $>$185 & \textbf{65.45 $\pm$ 26.48} \\
                \hline
                \end{tabular}
      }
\end{table}

An attribute representation $A \in \mathbb{B}^m$ are tabulated following the format presented in \cite{reiningAttributeRepresentationHuman2018}. Few subjects have the same attribute representations. For instance, from Tab.~\ref{table:attributes}, subject $10$ and $14$ of LARa dataset have the same attribute representation in $A_1$ as their soft-biometric features fall into the same range. Similarly, subject $12$ and $13$ have the same attribute representation in $A_2$ as their soft-biometric features fall into similar ranges. As a result, attribute representations cannot be used to identify a particular individual, rather, a group of individuals with similar soft-biometrics characteristics. 

The \textit{tCNN-IMU}$_{MLP}$ network was trained eight times by leaving out one subject and testing the network with the left-out subject. The network used epoch $10$, $Lr$=$10^{-4}$ and mB size=$100$. The experiments were conducted using the NNA with the \emph{Cosine} and \emph{PRM} similarities; however, the presented values are using the \emph{Cosine} similarity, as the \emph{PRM} showed the same performance. The average of eight iterations is presented. 

Tab.~\ref{Table: t1cvNN} presents the leave one out cross-validation (LOOCV) average performance of $A_{[1,2]}$ attribute representations of LARa- Mbientlab and Motionminers and $A$ for PAMAP2. For $A_1$ and $A_2$, the performance of Gender identity (Gender ID) and Height attributes outperforms that of Weight and Age. As Tab.~\ref{Table: t1cvNN} shows, in $A_2$, persons with $A_2^{height[>180]}$ shows a better accuracy for LARa-Mbientlab. However, the improved performance can be attributed to the low amount of variations from the soft-biometrics of the individuals found in the dataset. Hence, the soft-biometrics identification using attribute representation can be said to have potential, provided we have a larger and dedicated dataset. 
        
\section{Conclusions}

This work explored the possibility of person identification using motion information obtained from OBD data by training DNNs, such as \textit{tCNN-IMU}$_{MLP}$ and \textit{tCNN-IMU}$_{LSTM}$ using benchmark HAR datasets. Further, the impact of activities on the identification process was analysed. Finally, soft-biometrics identification using attribute representations was deployed. The soft-biometrics were modelled as attribute representations of the individual based on the LARa-Mbientlab dataset recording protocol.

Experiments on the impact of channel-normalisation on \textit{tCNN-IMU}$_{MLP}$ using LARa-Mbientlab pointed out that positional bias does not exceed the benefits of channel-normalisation. Based on the experimentation with LARa, OPP, and PAMAP2, person identification is not restricted to gait activity. The impact of activities showed that noise from instruments used could impede identification. Consistent with the research by \cite{elkaderWearableSensorsRecognizing2018}, dormant activities have high classification accuracy in the OPP. However, experiments on PAMAP2 negated the observation. 
    
Attribute representation with higher dimensionality was found to have better performance. Gender identity and height soft-biometrics consistently provided exceptional results. Evidently, the results of attribute representation are relatively poor. Consequently, a larger dataset with soft-biometrics, number of participants, and recording settings are necessary to facilitate attribute representation based transfer learning. This dataset should consider different recording settings for minimising the biases caused by having a single or very few recording set-ups, e.g., sensor position and configuration, per individual. In addition, careful documentation of participants and recording protocols is required. Furthermore, the application of the XAI method- LRP to identify the features that contribute to identity needs to be explored.
    
\section*{Acknowledgment}
\footnotesize{The work on this publication was supported by Deutsche Forschungsgemeinschaft (DFG) in the context of the project Fi799/10-2 ''Transfer Learning for Human Activity Recognition in Logistics''.}
    
\clearpage
%
%
\bibliographystyle{splncs04}
\bibliography{egbib}
\end{document}